\documentclass{article}

\usepackage[nonatbib,final]{neurips_2019_ml4ps}

\usepackage[utf8]{inputenc} 
\usepackage[T1]{fontenc}    
\usepackage{hyperref}       
\usepackage{url}            
\usepackage{booktabs}       
\usepackage{amsfonts}       
\usepackage{nicefrac}       
\usepackage{microtype}      

\usepackage[numbers,square]{natbib}
\usepackage{graphicx}
\usepackage{bm}
\usepackage{listings}
\usepackage{amsmath}
\usepackage{caption}
\usepackage[ruled,noend]{algorithm2e}
\usepackage{tabularx}
\usepackage{bm}
\title{Embedded Constrained Feature Construction for High-Energy Physics Data Classification}

\author{No\"elie Cherrier\textsuperscript{1,2}
\And
Maxime Defurne\textsuperscript{1}
\And 
Jean-Philippe Poli\textsuperscript{2}
\And 
Franck Sabati\'e\textsuperscript{1}\\
\and
\textsuperscript{1} Irfu, CEA, Universit\'e Paris-Saclay, 91191, Gif-sur-Yvette cedex, France.\\
\textsuperscript{2} CEA, LIST, 91191, Gif-sur-Yvette cedex, France.
}

\begin{document}

\maketitle

\begin{abstract}
Before any publication, data analysis of high-energy physics experiments must be validated. This validation is granted only if a perfect understanding of the data and the analysis process is demonstrated. Therefore, physicists prefer using transparent machine learning algorithms whose performances highly rely on the suitability of the provided input features.
To transform the feature space, feature construction aims at automatically generating new relevant features. Whereas most of previous works in this area perform the feature construction prior to the model training, we propose here a general framework to embed a feature construction technique adapted to the constraints of high-energy physics in the induction of tree-based models. Experiments on two high-energy physics datasets confirm that a significant gain is obtained on the classification scores, while limiting the number of built features. Since the features are built to be interpretable, the whole model is transparent and readable.
\end{abstract}

\section{Introduction}

The entire universe is made of elementary particles such as leptons and quarks as explained by the standard model. To test this model, particle physicists try to recover properties of composite particles from these of elementary particles. The Higgs mechanism~\cite{Higgs_1964} and the strong interaction~\cite{Roberts:2016vyn} are actively studied in particle collisions to understand the origin of mass.

The principle of a data analysis is always the same: detect and combine particles from collisions and use energy/momentum conservation to isolate the signals of interest. These signals are usually rare with several sources of background. Machine learning techniques can make a major contribution to such an analysis as long as they pass the mandatory interpretability requirement, sometimes at the expense of performances: any detail of an analysis must be explainable and understood.

Decision trees are very good candidates to meet this transparency requirement~\cite{gilpin_explaining_2018}. However it is also well known that the choice of features can greatly affect their performances~\cite{john_irrelevant_1994}.
Generally, a preprocessing step of feature engineering is performed manually based on domain expertise.
In particle physics, quantities related to energy/mass/momentum balances and highly dependent on the process of interest are derived from base variables. Although understandable and analyzable by construction, nothing guarantees that these quantities are optimized for the analysis of the process of interest. 

The field of feature construction (FC) aims at automating the feature engineering step. In particular, embedded FC permits to better adjust the built features to a local data discrimination problem during the model induction.
In this work, we aim at developing techniques of embedded FC with a special attention to the readability of the final model.
The global transparency of a model also requires the interpretability of the built features themselves, which is the second focus of our work.

We first review related work in the field, before presenting our method and conducting experiments on two HEP datasets from two different physics studies: the Higgs mechanism at CERN on one hand and the strong interaction at Jefferson Laboratory on the other hand.

\section{Related work}

Most of the automatic feature generation algorithms are actually preprocessing methods: new features are built from mathematical functions of the original ones and the training is performed in a subsequent step with the new features.

The literature in automatic FC is very abundant and a survey can be found in \cite{sondhi_feature_2009}. 
Genetic programming (GP) algorithm is a popular method in the FC field \cite{krawiec_genetic_2002,otero_genetic_2003,smith_genetic_2005,neshatian_genetic_2008}.
It consists in evolving a population of $N$-tree-like individuals through mutations and crossovers while selecting good individuals for the next generation.


In \cite{anonymous}, the authors propose a method to adapt a GP algorithm to handle the constraints of HEP.
Indeed, in HEP, a feature is interpretable if it resembles some of the physics laws ruling our universe. These physics laws connect variables carrying compatible units: for instance, no physics formula exhibits the raw sum of an angle and an energy. Therefore, to enforce the combination of compatible features, \citet{anonymous} constrain the GP algorithm with a grammar and a transition matrix. The resulting features are interpretable to physics experts and the classification score is improved compared to the unconstrained version of the algorithm.

On the other hand, the field of embedded FC has not been so much explored yet.
\citet{ekart_using_2003} are the first to use a GP algorithm to find the best splitting feature at each node in the induction of a decision tree.
\citet{hutchison_embedding_2012} embed a Monte Carlo search of features in several tree-based ensemble methods, building one feature at each node of the tree. 

\section{Embedded constrained feature construction method during tree induction}

\citet{anonymous} mainly use a machine learning algorithm to evaluate the candidate features during the evolution.
In this work, we propose to use the split criterion in the decision trees or ensemble methods as the fitness function in the constrained GP algorithm of \citet{anonymous}, thus improving the speed of the algorithm.
Computing a single information gain is indeed faster than training a whole decision tree.

Besides, we experimentally limit the number of built features per tree to support overall intelligibility of the model and also to prevent overfitting.
When the number of allowed constructions is restrained, the features are built from the root and level by level, going down as the tree is formed.

We propose hereafter a generic framework for embedding FC into classical tree induction algorithms.
The induction of the most commonly used tree classifiers (e.g. C4.5 \cite{Quinlan:1993:CPM:152181} and CART \cite{breiman2017classification}) is made by sequentially constructing their nodes. For each node, the problem is to separate the data into two subsets while optimizing a criterion which depends on the particular induction algorithm. Most algorithms perform a search among the existing features to find the best split according to this criterion. 
We modify this feature search step by the function \texttt{findBestSplit} described in Algorithm~\ref{alg:tree_induction_fc}.

\begin{algorithm}
\caption{Generic induction of a node of a decision tree with embedded FC}\label{alg:tree_induction_fc}
\DontPrintSemicolon
\SetKwInOut{Input}{Input}  
\Input{$data$ in the current node\\
$n_f$ the number of built features so far in the tree\\
$depth$ the depth of the current node\\
$N_{max}$ the maximum number of features to build in the tree
}
 \KwResult{the inducted node}
  \SetKwProg{Fn}{Function}{}{end}
 \SetKwFunction{FBestSplit}{findBestSplit}
 \SetKwFunction{FCriterion}{splittingCriterion}
 \SetKwFunction{FConstruct}{constructionCondition}
 \Fn{\FBestSplit{data, depth, $n_f$}}{
        \eIf{\FConstruct($n_f$, depth)}{
            build splittingFeature with \FCriterion as fitness function\;\label{l:criterion1}
            $n_f \leftarrow n_f + 1$\;
        }{
            \ForEach{feature \normalfont{in} data}{
                compute \FCriterion on \textit{feature}\;\label{l:criterion2}
            }
            splittingFeature $\leftarrow$ feature obtaining the best criterion\;
        }
        \lIf{splittingFeature \normalfont{does not satisfy specific requirements}\label{l:requirements}}{
            \KwRet null
        }
        splitted $\leftarrow$ split data along splittingFeature\;
        \KwRet splitted\;
        
  }
\end{algorithm}

At each node during the induction of a decision tree, if the \texttt{constructionCondition} in Equation~\eqref{eq:condition_for_fc} below is met with depth $d$ and number of built features so far $n_f$, with a maximal allowed number $N_{max}$ of feature constructions, we replace the classical search among existing features by a FC method.
\begin{equation}
\label{eq:condition_for_fc}
    d \le \log_2 (1 + N_{max}) \quad \text{and} \quad n_f < N_{max}.
\end{equation}
This condition is designed to ensure that the feature constructions will be performed in the first layers of the tree. 
The final feature obtained by the FC algorithm is used as the feature to split the data at the current node.

We use the C4.5~\cite{Quinlan:1993:CPM:152181}, adaptive boosting~\cite{adaboost} and gradient boosting~\cite{friedman2001} algorithms in our experiments.
The induction of a C4.5 decision tree is made by maximizing the information gain at each node. Adaptive boosting with decision trees only changes the weights of training examples. However, the gradient boosting classifier uses the mean squared error (MSE) as the splitting criterion since the global classification problem becomes a regression problem in the weak tree classifiers.
Feature construction is done by replacing the fitness function by the splitting criterion of the induction algorithm, computed using the considered candidate feature.

\section{Experiments}

We consider two HEP classification problems in this study.

\paragraph{DVCS dataset}
At Jefferson Laboratory, an electron beam scatters off protons at rest in the lab frame. The objective is to discriminate between the DVCS interaction whose final state is composed of an electron, a proton, and a photon, and the $\pi^0$ production event which has a similar final state, except that the photon is replaced by a $\pi^0$. The later immediately decays into two photons and one of them may not be detected, mimicking a DVCS event.
The available features are the three-dimensional momentum for each identified particle.

\paragraph{Higgs dataset \cite{adam-bourdarios_learning_2014}}
At CERN, two protons collide head-on with each other and Higgs particles are notably produced out of the collisions. 
The objective of this dataset is to detect Higgs bosons decaying into two $\tau$-particles.
The simulated data is publicly available on the Open Data platform of CERN. 
17 primitive features per event are available, including notably several geometrical features for each detected particle.

For both datasets, we use 100000 instances: 80\% of the them are dedicated to training and 20\% to performance evaluation. One should note that both datasets come from simulated data since the truth information is needed for training. Features already present in the datasets include the three-dimensional momenta of the particles as well as their $\theta$ and $\phi$ angles.
For the GP algorithm, we evolve a population of 500 individuals during 70 generations. For ensemble methods, we use a downgraded version of GP called ``GP down'' performing only 6 generations.

\subsection{Performance evaluation}

In this paragraph, we evaluate the performances of C4.5, AdaBoost and GradientBoosting classifiers with the embedded GP algorithm described above.

\begin{figure}
\centering
\includegraphics[width=\linewidth]{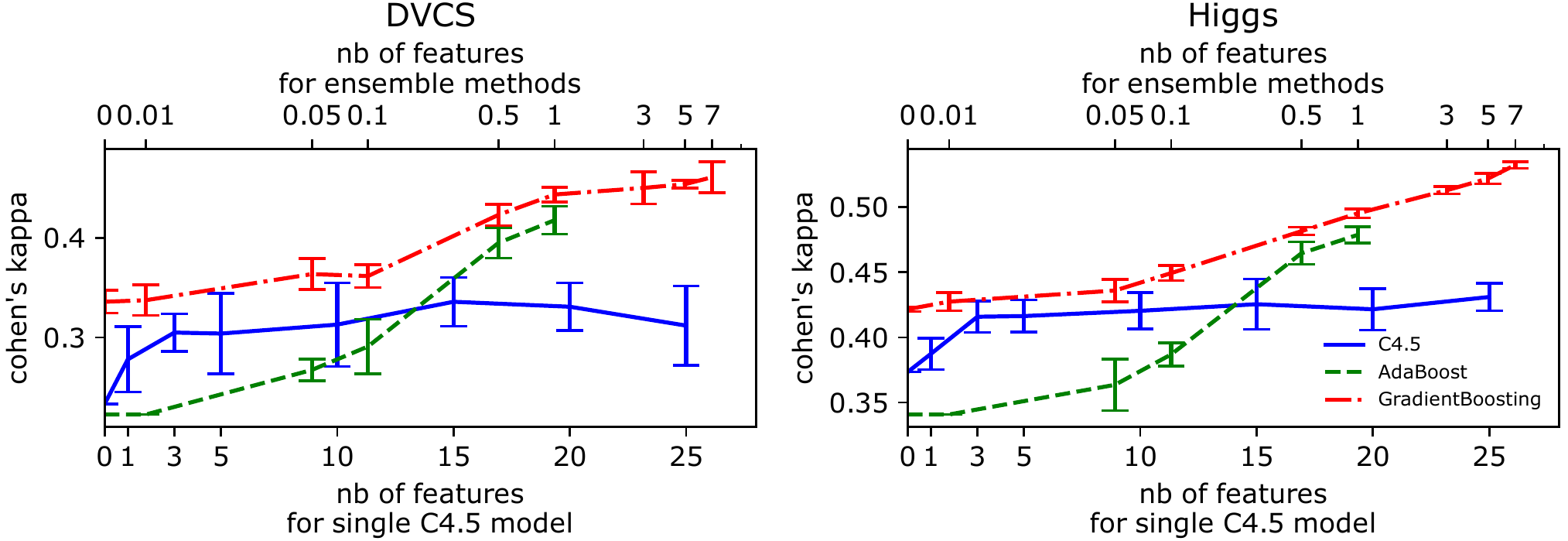}
\caption{Impact of the number of features on the classification score. The scale on the y-axis is different from one graph to the next.}
\label{fig:feature_nb}
\end{figure}

We vary the number of features built in total for C4.5 or per tree for ensemble methods. We use the Cohen's kappa metric to evaluate the performances, with at least 10 independent runs per displayed value.
Figure~\ref{fig:feature_nb} displays the evolution of the Cohen's kappa score of these three tree-based models depending on the number of built features.
For ensemble methods, we reserve the possibility to build less than one feature per tree in the ensemble: in the x-axis on the top of the graph, a value $p$ below 1 means that each tree in the ensemble has probability $p$ to build one feature. Above 1, the value $p$ can only be an integer and is the number of features built per tree.
The expectation for the total number of built features in the whole algorithm is then the product of this value $p$ with the number of trees in the ensemble.

The results show that the score increases with the number of built features per tree for ensemble methods. However, it stagnates for C4.5 algorithm after around 10-15 built features.
This shows that embedded FC with constrained GP brings a significant gain to the classification score compared to using the same tree-based algorithm without FC (score corresponding to zero built feature on Figure~\ref{fig:feature_nb}. Moreover, a reduction of the number of built features in C4.5 can be performed without impairing the score.

\begin{table}
\caption{Performance comparison (Cohen's kappa metric).\\
\textsuperscript{*} For DVCS, two hidden layers with respectively 100 and 50 neurons. For Higgs, one hidden layer with 150 neurons. This is the configuration leading to the best score after a grid search on up to three hidden layers and from 50 to 150 neurons per layer.}
\label{tab:my_label}
\centering
\begin{tabular}{lcc}
    \toprule
    & DVCS & Higgs \\
    \midrule
     C4.5 & 0.223 & 0.373 \\
     AdaBoost & 0.223 & 0.341 \\
     GradientBoosting & $0.336\pm0.011$ & $0.4216\pm0.0016$\\
    Neural network\textsuperscript{*} & $0.333\pm0.036$ & $0.337\pm0.062$\\
    Linear SVM & 0.343 & 0.169 \\
    \midrule
    FC C4.5 best & $0.3446\pm0.0080$ & $0.450\pm0.013$ \\
    FC AdaBoost best & $0.424\pm0.014$ & $0.4787\pm0.0062$ \\
    FC GradientBoosting best & $\bm{0.461\pm0.016}$ & $\bm{0.5325\pm0.0025}$ \\
    \bottomrule
\end{tabular}
\end{table}

Table~\ref{tab:my_label} provides a few baselines, including the three tree-based algorithms without FC, a neural network and a SVM. These baselines are compared to the best scores (i.e. the highest point on Figure~\ref{fig:feature_nb}) obtained with the three tree-based algorithms with embedded FC. Even compared to black-box models which are supposed to have an internal complex representation of the feature space, the three presented algorithms with embedded FC perform better.

\subsection{Model readability}

Model conciseness is supported by the smaller number of built features required by C4.5 to get an optimal classification score, and by the small size of weak tree classifiers which are limited in depth in the used implementations of the two presented ensemble methods (depth 1 for AdaBoost and 3 for GradientBoosting). The final feature space obtained with embedded FC is hence of reduced size. 

Equations~\eqref{eq:higgs1} and~\eqref{eq:higgs2} display two features built for the Higgs dataset that are either recurrent in their simple form or recurrent as a pattern in more complex features (they appear in respectively 61\% and 74\% of runs):

\begin{minipage}[c]{0.45\linewidth}
\begin{equation}
\label{eq:higgs1}
    \cos\left(\theta^{lep} - \theta^{\tau}\right)
\end{equation}
\end{minipage}
\begin{minipage}[c]{0.45\linewidth}
\begin{equation}
\label{eq:higgs2}
\cos\left(\phi^{lep} - \phi^{\tau}\right)
\end{equation}
\end{minipage}

These features visibly compares the geometrical angles of a lepton and a $\tau$, two particles which are indeed expected to be the products of the decay of the same particle. The mass of that particle can be computed from the features displayed above.

Equation~\eqref{eq:clas12} shows a recurrent feature built for the DVCS dataset (it appears in 79\% of runs):
\begin{equation}
\label{eq:clas12}
p_z^{e} + p_z^{\gamma_1} + p_z^{p}
\end{equation}
This feature is a momentum conservation check along the beam direction, absolutely relevant considering the detector and event geometries of a fixed-target experiment. A $\pi^0$ event would obviously miss a second photon momentum in the $p_z$ sum so this feature would not take the same value.

\section{Conclusion}

In the machine learning field, there is generally a need to strike a balance between the interpretability and the performances of a model.
In this work, without loss of generality, we improved three tree-based models by embedding a constrained FC technique adapted to physics problems during the induction.
In any case, embedded FC permitted to improve the classification score while experimenting on two datasets of HEP.
The constrained versions of the FC methods enables to build features that are interpretable at least to the experts of the field.

With the proposed method, the final model is more performant while remaining readable for further interpretation.
Finally, instead of choosing between performance and interpretability, we increased performance while focusing on keeping the readability of the final model.



\bibliographystyle{unsrtnat} 
\bibliography{main}

\begin{thebibliography}{17}
\providecommand{\natexlab}[1]{#1}
\providecommand{\url}[1]{\texttt{#1}}
\expandafter\ifx\csname urlstyle\endcsname\relax
  \providecommand{\doi}[1]{doi: #1}\else
  \providecommand{\doi}{doi: \begingroup \urlstyle{rm}\Url}\fi

\bibitem[Higgs(1964)]{Higgs_1964}
Peter~W. Higgs.
\newblock Broken symmetries and the masses of gauge bosons.
\newblock \emph{Phys. Rev. Lett.}, 13:\penalty0 508--509, Oct 1964.

\bibitem[Roberts(2017)]{Roberts:2016vyn}
Craig~D. Roberts.
\newblock {Perspective on the origin of hadron masses}.
\newblock \emph{Few Body Syst.}, 58\penalty0 (1):\penalty0 5, 2017.

\bibitem[Gilpin et~al.(2018)Gilpin, Bau, Yuan, Bajwa, Specter, and
  Kagal]{gilpin_explaining_2018}
Leilani~H. Gilpin, David Bau, Ben~Z. Yuan, Ayesha Bajwa, Michael Specter, and
  Lalana Kagal.
\newblock Explaining {Explanations}: {An} {Approach} to {Evaluating}
  {Interpretability} of {Machine} {Learning}.
\newblock \emph{arXiv:1806.00069 [cs, stat]}, May 2018.

\bibitem[John et~al.(1994)John, Kohavi, and Pfleger]{john_irrelevant_1994}
George~H. John, Ron Kohavi, and Karl Pfleger.
\newblock Irrelevant {Features} and the {Subset} {Selection} {Problem}.
\newblock In \emph{Machine {Learning} {Proceedings} 1994}, pages 121--129.
  Elsevier, 1994.

\bibitem[Sondhi(2009)]{sondhi_feature_2009}
Parikshit Sondhi.
\newblock Feature {Construction} {Methods}: {A} {Survey}.
\newblock page~8, 2009.

\bibitem[Krawiec(2002)]{krawiec_genetic_2002}
Krzysztof Krawiec.
\newblock Genetic {Programming}-based {Construction} of {Features} for
  {Machine} {Learning} and {Knowledge} {Discovery} {Tasks}.
\newblock page~15, 2002.

\bibitem[Otero et~al.(2003)Otero, Silva, Freitas, and
  Nievola]{otero_genetic_2003}
Fernando E.~B. Otero, Monique M.~S. Silva, Alex~A. Freitas, and Julio~C.
  Nievola.
\newblock Genetic {Programming} for {Attribute} {Construction} in {Data}
  {Mining}.
\newblock In \emph{Proceedings of the 6th {European} {Conference} on {Genetic}
  {Programming}}, {EuroGP}'03, pages 384--393, Berlin, Heidelberg, 2003.
  Springer-Verlag.
\newblock ISBN 978-3-540-00971-9.

\bibitem[Smith and Bull(2005)]{smith_genetic_2005}
Matthew~G. Smith and Larry Bull.
\newblock Genetic {Programming} with a {Genetic} {Algorithm} for {Feature}
  {Construction} and {Selection}.
\newblock \emph{Genetic Programming and Evolvable Machines}, 6\penalty0
  (3):\penalty0 265--281, September 2005.

\bibitem[Neshatian and Zhang(2008)]{neshatian_genetic_2008}
Kourosh Neshatian and Mengjie Zhang.
\newblock Genetic {Programming} and {Class}-{Wise} {Orthogonal}
  {Transformation} for {Dimension} {Reduction} in {Classification} {Problems}.
\newblock In \emph{Genetic {Programming}}, Lecture {Notes} in {Computer}
  {Science}, pages 242--253. Springer Berlin Heidelberg, 2008.
\newblock ISBN 978-3-540-78671-9.

\bibitem[{Cherrier} et~al.(2019){Cherrier}, {Poli}, {Defurne}, and
  {Sabatié}]{anonymous}
N.~{Cherrier}, J.~{Poli}, M.~{Defurne}, and F.~{Sabatié}.
\newblock {Consistent Feature Construction with Constrained Genetic Programming
  for Experimental Physics}.
\newblock In \emph{2019 IEEE Congress on Evolutionary Computation (CEC)}, pages
  1650--1658, June 2019.

\bibitem[Ek\'art and M\'arkus(2003)]{ekart_using_2003}
Anik\'o Ek\'art and Andr\'as M\'arkus.
\newblock Using genetic programming and decision trees for generating
  structural descriptions of four bar mechanisms.
\newblock \emph{AI EDAM}, 17\penalty0 (03), August 2003.

\bibitem[Maes et~al.(2012)Maes, Geurts, and Wehenkel]{hutchison_embedding_2012}
Francis Maes, Pierre Geurts, and Louis Wehenkel.
\newblock Embedding {Monte} {Carlo} {Search} of {Features} in {Tree}-{Based}
  {Ensemble} {Methods}.
\newblock In \emph{Machine {Learning} and {Knowledge} {Discovery} in
  {Databases}}, volume 7523, pages 191--206. Springer Berlin Heidelberg,
  Berlin, Heidelberg, 2012.

\bibitem[Quinlan(1993)]{Quinlan:1993:CPM:152181}
J.~Ross Quinlan.
\newblock \emph{C4.5: Programs for Machine Learning}.
\newblock Morgan Kaufmann Publishers Inc., San Francisco, CA, USA, 1993.

\bibitem[Breiman(2017)]{breiman2017classification}
Leo Breiman.
\newblock \emph{Classification and regression trees}.
\newblock Routledge, 2017.

\bibitem[Zhu et~al.(2006)Zhu, Rosset, Zou, and Hastie]{adaboost}
Ji~Zhu, Saharon Rosset, Hui Zou, and Trevor Hastie.
\newblock Multi-class adaboost.
\newblock \emph{Statistics and its interface}, 2, 02 2006.

\bibitem[Friedman(2001)]{friedman2001}
Jerome~H. Friedman.
\newblock Greedy function approximation: A gradient boosting machine.
\newblock \emph{Ann. Statist.}, 29\penalty0 (5):\penalty0 1189--1232, 10 2001.

\bibitem[Adam-Bourdarios et~al.(2014)Adam-Bourdarios, Cowan, Germain, Guyon,
  Kegl, and Rousseau]{adam-bourdarios_learning_2014}
Claire Adam-Bourdarios, Glen Cowan, Cecile Germain, Isabelle Guyon, Balazs
  Kegl, and David Rousseau.
\newblock Learning to discover: the {Higgs} boson machine learning challenge -
  {Documentation}.
\newblock 2014.

\end{thebibliography}

\end{document}